\documentclass[letterpaper, 10 pt, conference]{ieeeconf}  % Comment this line out if you need a4paper

\IEEEoverridecommandlockouts                              % This command is only needed if 
\usepackage{graphics} % for pdf, bitmapped graphics files
\usepackage{epsfig} % for postscript graphics files
\usepackage{mathptmx} % assumes new font selection scheme installed
\usepackage{times} % assumes new font selection scheme installed
\usepackage{amsmath} % assumes amsmath package installed
\usepackage{amssymb}  % assumes amsmath package installed

\usepackage{multirow}   % for \multirow
\usepackage{booktabs}   % for \toprule, \midrule, \bottomrule, \cmidrule
\usepackage{graphicx}   % for \resizebox
\usepackage{subcaption}
\usepackage{balance}
\usepackage{cite}

\usepackage[table]{xcolor}
\usepackage{soul}
\DeclareMathAlphabet{\mathcal}{OMS}{cmsy}{m}{n}
\newcommand{\methodname}{\text{FedDRMan}}

\usepackage[pagebackref=true,breaklinks=true,letterpaper=true,colorlinks,bookmarks=false]{hyperref}

\title{\LARGE \bf
\methodname{}: Federated Subspace Guided Vision-Language-Action Policy Distillation for Non-IID Multi-Robot Manipulation
}

\author{Biprodip Pal$^1$,
Kaushik Roy$^2$,
Yanming Zhu$^1$, 
Brendan Tidd$^2$, 
Alan Wee-Chung Liew$^{1*}$,
Peyman Moghadam$^2$ % <-this % stops a space
\thanks{$^1$ School of ICT, Griffith University, Australia}
\thanks{$^2$ CSIRO Robotics, CSIRO, Australia}\protect\\
\thanks{{\tt\small$^*$Corresponding author:a.liew@griffith.edu.au}}
}

\begin{document}

\bstctlcite{IEEEexample:BSTcontrol}

\maketitle
\thispagestyle{empty}
\pagestyle{empty}

\begin{abstract}

Federated learning offers a natural way for multiple robots to jointly improve manipulation policies without requiring centralized access to training demonstrations. However, non-IID task and environment distributions can induce representation drift and mutually incompatible robot-policy updates, making naive parameter aggregation destructive. We present FedDRMan, a federated subspace-guided distillation framework for heterogeneous robot manipulation. At each communication round, the server model provides a frozen teacher for local behavior cloning, while low-rank multimodal subspace and action-distribution distillation preserve globally useful representation geometry and policy behavior. To address heterogeneous aggregation, FedDRMan groups clients by update compatibility and maintains a persistent model for each cluster. The server then spectrally rebalances each compatible aggregate to mitigate attenuation of weaker task-relevant robot-policy update directions. Extensive experiments on LIBERO across diverse non-IID settings, heterogeneity levels, client participation variation, together with ablations and aggregation analyses, show that FedDRMan substantially improves knowledge transfer and consistently outperforms strong federated baselines achieving a peak mean success rate of $80.7\%$, $11.6$ percentage points above the strongest evaluated federated baseline.

\end{abstract}

\section{Introduction}
\label{sec:introduction}

Vision-Language-Action (VLA) policies have shown strong progress in
language-conditioned robot manipulation by combining visual perception,
language understanding, and low-level control. Systems such as RT-2 and
OpenVLA-OFT demonstrate strong performance in VLA-based manipulation and
robotic control~\cite{brohan2023rt2,kim2025oft}. This progress often depends
on large and diverse robot datasets. For example, Open X-Embodiment combines
demonstrations from multiple robot platforms, tasks, and environments to
support cross-robot generalization~\cite{oneill2024openx}. However, most VLA
training and adaptation methods still assume that demonstrations from
different robots can be pooled for centralized training. This assumption
limits multi-robot systems, where repeatedly transferring raw trajectories to
a central server can be communication-intensive and may reveal sensitive
information about users, robot locations, operating environments, or other
deployment details. Collaborative VLA training therefore requires mechanisms that exploit distributed robot experience without centralizing the underlying demonstrations. 

Federated learning (FL) provides a natural foundation for multi-robot
collaboration: each robot learns from its local demonstrations and exchanges
model updates rather than raw data, allowing distributed knowledge to
contribute to a shared policy~\cite{mcmahan2017communication}. However,
heterogeneous local data can cause robot-policy drift and ineffective aggregation
~\cite{li2020fedprox}. For example, one robot may observe mostly
object-centric pick-and-place tasks, while another sees goal-conditioned tasks
in different scenes and viewpoints, producing different multimodal feature
distributions and update directions. 
Local behavior cloning (BC) can therefore shift visual, language, and proprioceptive representations toward client-specific feature distributions, while aggregating incompatible updates may suppress task-relevant knowledge.
%Local behavior cloning (BC) can therefore drive visual, language, and proprioceptive representations toward client-robot-specific distributions, while aggregating incompatible updates may suppress task-relevant knowledge. 
Much existing work on heterogeneous FL,
however, has focused primarily on unimodal or non-robotic settings
~\cite{dFedDG2,li2026resource,FedKD}.

Knowledge distillation provides a mechanism for constraining local policy drift by transferring knowledge across tasks and models in federated and continual learning settings~\cite{roy2025m2distill,FedKD}.
%Knowledge distillation offers an effective mechanism for transferring knowledge across tasks and models in federated and continual learning settings. 
Recent work further highlights the importance of representation geometry and subspace structure for knowledge transfer. Subspace-based distillation reduces the sensitivity of coordinate-wise feature matching in high-dimensional representations~\cite{roy2026spread}, while spectral model-merging methods identify task-relevant subspaces and preserve specialized update directions~\cite{essential_subspace_merging,skorobogat2025subspace}.

However, these techniques have mainly been developed for sequential learning or one-shot centralized model merging, rather than federated VLA training, where repeated local adaptation reshapes multimodal representations and heterogeneous robot-policy updates must be periodically integrated.
%However, these techniques are mainly developed for sequential learning or one-shot centralized model merging, rather than federated VLA training, where non-IID clients continuously reshape multimodal policies and heterogeneousupdates are periodically aggregated. 
Recent methods such as FLAME~\cite{betran2025flame} and FedVLA~\cite{miao2025fedvla} demonstrate the feasibility of federated robotic manipulation. However, the coupled effects of multimodal representation drift during local adaptation and incompatible client updates during server aggregation remain insufficiently studied under task and environment heterogeneity.

We introduce \textbf{FedDRMan}, \emph{\textbf{Fed}erated
Subspace Guided Vision-Language-Action Policy \textbf{D}istillation 
for Non-IID Multi-\textbf{R}obot \textbf{Man}ipulation}. FedDRMan performs
geometry-aware global-to-local knowledge transfer by distilling multimodal
representation subspaces and action distributions from a compatible server
policy during local adaptation. At the server, FedDRMan exploits client-update geometry to separate incompatible robot-policy updates and redistribute the strength of update directions within each compatible aggregate. Together, these mechanisms preserve transferable policy structure during local adaptation and heterogeneous knowledge integration at the server, without additional communication. Our contributions are threefold:

\begin{itemize}
    \item We formulate federated multimodal subspace
    distillation for non-IID manipulation, regularizing
    representation geometry and action distributions against a frozen server policy during local behavior cloning.

    \item We introduce compatibility-aware clustered spectral aggregation for heterogeneous federation, grouping incompatible robot-policy
    updates before full-model aggregation and spectrally rebalancing each compatible aggregate to preserve weaker task-relevant directions
    without additional communication.

   \item We demonstrate strong and consistent gains across multiple LIBERO
   non-IID settings, including same-suite and cross-suite federation and varying degrees of task and environment heterogeneity, together with extensive ablation and aggregation analyses.
\end{itemize}

\section{Related Work}
\label{sec:related_work}

% \paragraph{Federated Robot Learning under Heterogeneity}
Prior non-IID FL methods mitigate heterogeneity through proximal
regularization, distribution-aware collaboration, or decoupled
personalization
~\cite{li2020fedprox,dFedDG2,collins2021fedrep}, but are not tailored
to multimodal robot manipulation policies. Federated robotics has previously been studied for object grasping,
swarm navigation, multi-robot spatio-temporal prediction, and warehouse
task scheduling~\cite{kang2023fogl,na2023federatednavigation,
majcherczyk2021flowfl,ho2024federatedtask}. Federated reinforcement
learning has enabled robots to collaboratively learn navigation policies,
while Flow-FL studies federated spatio-temporal prediction from
continuously collected multi-robot data~\cite{ho2024federatedtask,majcherczyk2021flowfl}. However,
their learning objectives remain task-specific and generally do not
address end-to-end multimodal, language-conditioned manipulation or
representation drift across heterogeneous robot policies. FLAME subsequently established a federated manipulation benchmark and
showed that standard FL baselines can struggle on harder manipulation
tasks, but is primarily a benchmark rather than a
heterogeneity-specific learning method~\cite{betran2025flame}.
FedVLA directly addresses federated VLA learning through
instruction-oriented scene parsing, a dual-gating mixture-of-experts
(MoE), and expert-driven aggregation~\cite{miao2025fedvla}. Its
heterogeneity handling is coupled to MoE expert activation and
expert-specific aggregation, whereas FedDRMan targets multimodal
representation drift and incompatible client-update geometry without
requiring an expert-routed policy architecture.

% These methods are not designed specifically around multimodal manipulation-policy geometry.
% These works demonstrate that distributed robot experience can be exploited without centralizing raw data; for example, prior methods consider collaborative grasp learning, federated reinforcement learning for navigation, shared spatio-temporal prediction, and distributed task allocation.

% \subsection{Geometric Distillation and Model Aggregation}

Representation-preserving distillation has been studied in lifelong
robot learning, including multimodal feature alignment in M2Distill and
low-rank subspace preservation in SPREAD
\cite{roy2025m2distill,roy2026spread}. Related model-merging work
studies task-relevant spectral structure and preservation of weaker
update directions
\cite{essential_subspace_merging,skorobogat2025subspace}.
These methods primarily target sequential learning or one-shot approach. FedDRMan builds on these geometric principles but addresses a
distinct federated problem, jointly controlling multimodal
representation drift during local adaptation and destructive
interference during aggregation through server-guided subspace
distillation, compatibility-aware clustering, and spectral
rebalancing.

% \subsection{Geometry-Aware Distillation and Aggregation}
% Multimodal distillation has also been studied in lifelong robot learning. M2Distill aligns modality-specific representations between consecutive policies, while SPREAD replaces raw feature matching with SVD-based low-rank subspace alignment to preserve representation geometry during sequential adaptation~\cite{roy2025m2distill,roy2026spread}. FedDRMan adapts this principle to a federated global-to-local relation, using the round-level server policy as the teacher for non-IID clients. At the server, complementary model-merging work studies the spectral structure of heterogeneous updates: Essential Subspace Merging identifies task-relevant subspaces, while Subspace-Boosted Model Merging strengthens weaker singular directions through spectral redistribution~\cite{essential_subspace_merging,skorobogat2025subspace}. FedDRMan adapts the latter operation to repeated within-cluster federated aggregation, yielding a two-level geometric design in which subspace distillation constrains local representation drift and compatibility-aware spectral aggregation structures global knowledge integration.

\section{Method}
\label{sec:method}

\begin{figure*}[thpb]
    \centering
    \includegraphics[width=0.95\textwidth]{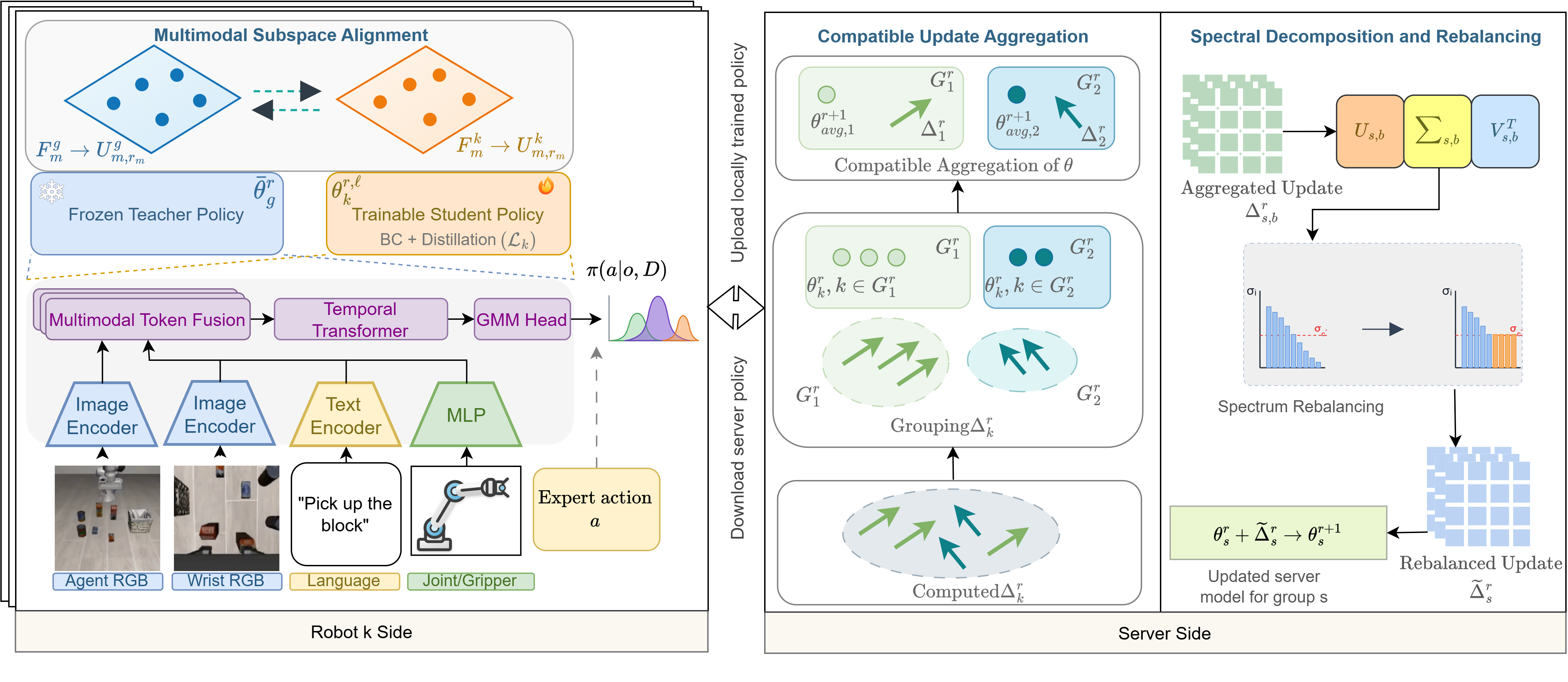}
    \caption{
Overview of FedDRMan. Multimodal subspace distillation limits
representation drift during local adaptation, while compatible-update
aggregation and spectral rebalancing mitigates destructive aggregation of
conflicting non-IID robot-policy updates and preserve task-relevant
update directions.
    }
    \label{fig:feddrman_overview}
    \vspace{-1.0em}
\end{figure*}

\subsection{Problem Formulation}
\label{subsec:problem}

We consider a federation of $K$ robots operating in a shared language-conditioned manipulation setting 
formulated as a finite-horizon Markov decision process (MDP)
$\mathcal{M}=(\mathcal{S},\mathcal{A},\mathcal{T},H,\mu_0,\phi)$, where
$\mathcal{S}$ is the state space, $\mathcal{A}$ is the continuous action space, $\mathcal{T}:\mathcal{S}\times\mathcal{A}\rightarrow\mathcal{S}$
denotes the transition dynamics, $H$ is the episode horizon, $\mu_0$ is
the initial-state distribution, and $\phi:\mathcal{S}\rightarrow\{0,1\}$
is the sparse goal predicate indicating task success during closed-loop evaluation. At timestep $t$, the robot receives a multimodal observation $o_t$ of the underlying state $s_t\in\mathcal{S}$, executes an action $a_t\in\mathcal{A}$, and transitions to $s_{t+1}=\mathcal{T}(s_t,a_t)$. Each client $k$ interacts with a local task distribution and owns a private demonstration set
\begin{equation}
    \mathcal{D}_k
    = \{\tau_i^k\}_{i=1}^{n_k},
    \qquad
    \tau_i^k
    = \{(o_{i,t}^k,D_i^k,a_{i,t}^k)\}_{t=1}^{T_i},
    \quad T_i \le H.
\end{equation}
where $D_i^k$ is the language instruction, $a_{i,t}^k\in\mathcal{A}$ is the expert action, and $o_{i,t}^k$ contains agent-view RGB, wrist RGB, joint state, and gripper state. Following LIBERO~\cite{liu2023libero}, each trajectory is collected from an expert policy and provides supervision for a language-conditioned Gaussian mixture model (GMM) policy $\pi_{\theta}(a\mid o,D)$.

The local distributions $\mathcal{D}_k$ are assumed to be non-IID: different robots may observe different task families, demonstrations, and state-action distributions. Let $\theta_g^r$ denote the server model at communication round $r$. Each participating client initializes its local policy from $\theta_g^r$ and optimizes using non-shared local $\mathcal{D}_k$. The overall federated objective is,

\begin{equation}
    \min_{\theta}
    \sum_{k=1}^{K} p_k
    \,\mathbb{E}_{(o,D,a)\sim\mathcal{D}_k}
    [\mathcal{L}_k(\theta;o,D,a)],
    \qquad
    p_k=\frac{n_k}{\sum_j n_j},
    \label{eq:fed_objective}
\end{equation}

Under non-IID $\mathcal{D}_k$, independent local adaptation can induce robot-specific representation and policy drift, such that direct parameter averaging may yield a global policy that poorly preserves behaviors across heterogeneous tasks and environments. FedDRMan therefore regularizes each local policy against the server model to preserve modality-specific
subspace geometry, while structuring how heterogeneous updates are subsequently merged.

% subject to all demonstrations remaining local. Under heterogeneous $\mathcal{D}_k$, independent local adaptation can induce representation and policy drift before aggregation; FedDRMan therefore regularizes each local policy against the current server model while structuring how heterogeneous updates are subsequently merged. 
% When the proposed distillation and spectral aggregation are disabled, the formulation reduces to FedAvg~\cite{mcmahan2017communication}.

\subsection{Federated Subspace-Guided Global-to-Local Distillation}
\label{subsec:fed_distill}

Figure~\ref{fig:feddrman_overview} illustrates our approach. At round $r$, let $\theta_g^r$ denote the server model assigned to client $k$.
Client $k$ creates a frozen teacher $\bar\theta_g^r$ and a trainable
student $\theta_k^{r,0}$ from this model.
The teacher remains fixed during the local updates of round $r$ and is
refreshed after server aggregation:
\begin{equation}
    \bar\theta_g^r\leftarrow\theta_g^r,\qquad
    \theta_k^{r,0}\leftarrow\theta_g^r,\qquad
    \nabla_{\bar\theta_g^r}=0.
    \label{eq:teacher}
\end{equation}

Each client then optimizes the local policy on its private demonstrations
using BC, regularized by representation and action-level distillation
terms. For readability, we write $\theta_k$ for the current local student
parameters $\theta_k^{r,\ell}$ within the local loss terms. Local
adaptation to the expert demonstrations is driven by the standard BC
objective
\begin{equation}
    \mathcal{L}_{\mathrm{BC}}(\theta_k)
    =-\mathbb{E}_{(o,D,a)\sim\mathcal{D}_k}
    [\log\pi_{\theta_k}(a\mid o,D)].
    \label{eq:bc}
\end{equation}

Under non-IID local adaptation, BC can reshape modality-specific
feature geometry away from the shared server representation.
Motivated by representation-preserving distillation~\cite{FedKD,roy2026spread},
we preserve the dominant representation subspaces of the server policy.
The leading singular directions capture principal representation
structure, allowing shared task-relevant geometry to be constrained
while retaining lower-energy directions for robot-specific adaptation.

For modality $m\in\mathcal{M}$, with
$\mathcal{M}=\{\text{text},\text{agent},\text{wrist},\text{joint},
\text{gripper}\}$, the batch-time encoder features are stacked into
$F_m^x\in\mathbb{R}^{d_m\times N}$, $N=BT$, where
$x\in\{g,k\}$ denotes the server teacher or local student, respectively.
The reduced singular value decomposition (SVD) and rank-$r_m$
orthogonal projector are
\begin{equation}
    F_m^x=U_m^x\Sigma_m^x(V_m^x)^\top,\qquad
    P_{r_m}^x=U_{m,r_m}^x(U_{m,r_m}^x)^{\top}.
    \label{eq:subspace_projection}
\end{equation}
The projected representation is therefore $P_{r_m}^xF_m^x$.
A small pre-SVD perturbation is applied for numerical stability when
the feature matrix is nearly degenerate.

We then define a symmetric subspace distillation loss for each modality
$m$:
\begin{align}
\mathcal{L}_{\mathrm{sub}}^m
&=
\|P_{r_m}^gF_m^g-P_{r_m}^kF_m^k\|_F^2
+
\|P_{r_m}^gF_m^k-P_{r_m}^kF_m^g\|_F^2.
\label{eq:subspace_loss}
\end{align}

The first term aligns feature content retained by the teacher and
student subspaces, while the second enforces symmetric cross-projection
consistency. Together, they preserve dominant representation geometry
without requiring coordinate-wise feature correspondence.

% Together, these terms encourage compatibility of the dominant
% representation geometry without requiring coordinate-wise
% correspondence. Consequently, the objective constrains the high-energy
% structure inherited from the server policy while retaining degrees of
% freedom in lower-energy directions, providing a principled balance
% between server representation consistency and client-specific
% plasticity.

Representation alignment alone does not guarantee consistent control
behavior. We therefore retain the teacher's action distribution by
sampling
$\hat a\sim\pi_{\bar\theta_g^r}(\cdot\mid o,D)$ and using
\begin{equation}
    \mathcal{L}_{\mathrm{act}}
    =
    \mathbb{E}_{\hat a\sim\pi_{\bar\theta_g^r}}
    \!\left[\log\pi_{\bar\theta_g^r}(\hat a\mid o,D)
    -
    \log\pi_{\theta_k}(\hat a\mid o,D)
    \right].
    \label{eq:action_distill}
\end{equation}

Log probabilities are clamped for numerical stability. The resulting
objective balances adaptation to local demonstrations with preservation
of useful server representation and policy structure:
\begin{equation}
    \mathcal{L}_k
    =
    \mathcal{L}_{\mathrm{BC}}
    +
    \sum_{m\in\mathcal{M}}
    \lambda_m \mathcal{L}_{\mathrm{sub}}^m
    +
    \lambda_{\mathrm{act}}\mathcal{L}_{\mathrm{act}}.
    \label{eq:local_objective}
\end{equation}

For local optimization step $\ell$, the student parameters are updated as
\begin{equation}
    \theta_k^{r,\ell+1}
    =
    \theta_k^{r,\ell}
    -
    \eta\nabla_{\theta_k^{r,\ell}}\mathcal{L}_k.
    \label{eq:local_update}
\end{equation}
Gradients are propagated only through the local student, while the
server teacher is held fixed. After the final local optimization step
$\ell_{\mathrm{final}}$, we denote the locally trained model uploaded
to the server by
$\theta_k^r= \theta_k^{r,\ell_{\mathrm{final}}}$.
Thus, BC drives robot-specific adaptation, subspace distillation
constrains representation drift, and action distillation preserves the
behavioral policy inherited from the server model.

\subsection{Compatible Spectral Federated Aggregation}
\label{subsec:aggregation}

After local training, robots upload their locally trained models to the
server. FedDRMan first groups clients by update compatibility, aggregates complete models within each cluster, and spectrally rebalances the resulting cluster updates. This prevents incompatible non-IID updates from being directly
aggregated and helps preserve task-relevant update directions.

\subsubsection{Compatible Update Aggregation (CUA)}
\label{subsubsec:cluster_aggregation}

Cross-suite heterogeneity in tasks, environments, and multimodal
observation distributions can induce incompatible robot-policy update
directions, making a single global consensus restrictive. FedDRMan
therefore maintains $C$ persistent cluster models
$\{\theta_s^r\}_{s=1}^{C}$ at the server, while keeping the local
objective in Eq.~\eqref{eq:local_objective} unchanged. All clients
share the common initialization $\theta^0$. Compatibility between robot-policy updates is characterized by the accumulated update from the common initialization,
\begin{equation}
    \Delta_k^r
    =
    \theta_k^r-\theta^0,
    \qquad
    v_k^r
    =
    \operatorname{vec}
    \!\left(
    \left.\Delta_k^r\right|_{\Omega}
    \right),
    \qquad
    S_{ij}^r
    =
    \frac{(v_i^r)^\top v_j^r}
         {\|v_i^r\|_2\|v_j^r\|_2},
    \label{eq:cluster_similarity}
\end{equation}

where $\Omega$ denotes a selected shared submodule of the policy and
$S_{ij}^r$ is the similarity between robot-policy update
representations. The accumulated update is used only for clustering,
as it provides a more stable compatibility signal than the diminishing
incremental per-round update, as analyzed in
Sec.~\ref{subsec:aggregation_effectiveness}. Complete local models are
used for aggregation.

% where $\Omega$ denotes a selected shared submodule of the policy, and
% $S_{ij}^r$ is the cosine similarity between the accumulated update
% representations of clients $i$ and $j$. The restricted accumulated
% update is used only for clustering. We use the from-initial
% representation because incremental per-round updates decrease in
% magnitude as training progresses and may provide a less stable
% compatibility signal, as analyzed in
% Sec.~\ref{subsec:aggregation_effectiveness}. Aggregation itself uses
% the complete locally trained model.

Let $\overline{S}(G,G')$ denote the mean of $S_{ij}^r$ over distinct
client pairs $i\in G$, $j\in G'$, with $i\neq j$. Clients are partitioned into $C$
non-empty groups by maximizing mean intra-cluster similarity while
minimizing mean inter-cluster similarity:
\small
\begin{equation}
    \{G_s^r\}_{s=1}^{C}
    =
    \operatorname*{arg\,max}_{\{G_s\}\in\Pi_C}
    \left[
    \frac{1}{C}\sum_s\overline{S}(G_s,G_s)
    -
    \binom{C}{2}^{-1}
    \sum_{s<s'}\overline{S}(G_s,G_{s'})
    \right],
    \label{eq:cluster_partition}
\end{equation}
\normalsize

where $G_s^r\subseteq\{1,\ldots,K\}$ denotes the set of clients assigned to cluster $s$ at round $r$, and $\Pi_C$ denotes the set of all partitions into exactly $C$ non-empty groups.

Within cluster $s$, the sample weights are renormalized as
$\widetilde{p}_k = \frac{n_k}{\sum_{j\in G_s^r} n_j}$,
and the complete local models are aggregated:
\begin{equation}
    \theta_{\mathrm{avg},s}^{r+1}
    =
    \sum_{k\in G_s^r}
    \widetilde{p}_k\theta_k^r,
    \qquad
    \Delta_s^r
    =
    \theta_{\mathrm{avg},s}^{r+1}-\theta_s^r.
    \label{eq:avg_delta}
\end{equation}
% Thus, $\Delta_s^r$ is the cluster-level update relative to the corresponding persistent server model.

\subsubsection{Spectral Rebalancing (SRB)}
\label{subsubsec:spectral_rebalancing}

% Direct averaging can attenuate weaker but potentially useful directions
% in $\Delta_s^r$. FedDRMan therefore spectrally rebalances each eligible floating-point
% two-dimensional weight matrix in the cluster update. While prior work ~\cite{skorobogat2025subspace,essential_subspace_merging},
% studied such spectral operations in one-shot centralized model merging,
% we use SRB conservatively as a spectral redistribution mechanism for
% round-wise federated updates and do not explicitly assume or measure
% rank collapse in our setting.

Direct averaging can attenuate weaker but useful directions in
$\Delta_s^r$. FedDRMan applies SRB to eligible two-dimensional weight matrices, motivated by spectral model-merging methods
~\cite{skorobogat2025subspace,essential_subspace_merging}. Unlike the original
one-shot merging setting, here, SRB serves as a spectral redistribution mechanism for federated updates.

Let $\Delta_{s,b}^r$ denote an eligible two-dimensional parameter block
$b$ of $\Delta_s^r$. Its singular value decomposition is
\begin{equation}
    \Delta_{s,b}^r
    =
    U_{s,b}\Sigma_{s,b}V_{s,b}^{\top},
    \qquad
    \Sigma_{s,b}
    =
    \operatorname{diag}(\sigma_1,\ldots,\sigma_R),
    \label{eq:srb_svd}
\end{equation}
where the round superscript on the SVD factors is omitted for
readability. The cumulative singular-value threshold
$\beta\in[0,1]$ determines
\begin{equation}
    c^\star
    =
    \min\left\{
    i:
    \frac{\sum_{j=1}^{i}\sigma_j}
         {\sum_{j=1}^{R}\sigma_j}
    \ge\beta
    \right\},
    \qquad
    \widetilde{\sigma}_i
    =
    \max(\sigma_i,\sigma_{c^\star}).
    \label{eq:boost}
\end{equation}
The rebalanced matrix update is then
$   \widetilde{\Delta}_{s,b}^r = U_{s,b}\widetilde{\Sigma}_{s,b}V_{s,b}^{\top}
$. Parameter blocks that are not eligible for
SRB retain their original within-cluster averaged updates. The next
persistent cluster model is therefore
$\theta_s^{r+1}
    =   \theta_s^r+\widetilde{\Delta}_s^r.
$

% SRB is applied to eligible linear weight matrices; all remaining
% parameters retain their within-cluster averaged updates. Clustering and
% SRB are server-side and therefore introduce no additional communication
% payload over FedAvg.

In practice, SRB is applied to eligible linear weight matrices in the
image spatial projector, language and proprioceptive encoders, temporal
transformer, and policy head; all remaining parameters retain their
within-cluster averaged updates. The server tracks client-to-cluster
assignments across rounds. At the beginning of round $r+1$, each client
downloads the server model associated with its assigned cluster to
initialize local training. Both clustering and SRB are performed entirely on the server and
introduce no additional client-server model payload relative to FedAvg.

% Biases, normalization parameters, convolutional kernels, and other
% non-matrix tensors are not spectrally modified and retain their standard within-cluster averaged updates.

% The additional server-side cost arises from the spectral decompositions
% of eligible matrices and from maintaining $C$ persistent cluster models.

% Boosted set is the Linear layer weight matrices throughout the policy including, Image-encoder's spatial projection layer,Language encoder,Proprio encoders. Temporal transformer (the submodule clustering itself is computed on), Policy head (the GMM head's linear layers).

% Hence, clustering determines \emph{which clients are merged}, while spectral rebalancing determines \emph{how the resulting cluster update is merged}.

\subsubsection{Theoretical Interpretation of FedDRMan Aggregation}
\label{subsubsec:heterogeneous_analysis}

Prior model-merging analysis shows that averaging heterogeneous
task-specific directions can attenuate unique information relative to
shared directions~\cite{skorobogat2025subspace}. A similar effect arises
in weighted federated aggregation. For $k\in G_s^r$, write
\begin{equation}
    \delta_k^r
    =
    \theta_k^r-\theta_s^r
    =
    C_s^r+U_k^r,
    \qquad
    \Delta_s^r
    =
    \sum_{k\in G_s^r}\widetilde p_k\delta_k^r,
    \label{eq:hetero_decomp}
\end{equation}
where $C_s^r$ is a component shared within cluster $s$ and $U_k^r$
captures robot-specific directions. Since
$\sum_k\widetilde p_k=1$, the unique component of the aggregate is
$\sum_k\widetilde p_kU_k^r$. If the $U_k^r$ are pairwise orthogonal
and $\|U_k^r\|_F\le B$, then
\begin{equation}
    \left\|
    \sum_k\widetilde p_kU_k^r
    \right\|_F^2
    =
    \sum_k\widetilde p_k^2\|U_k^r\|_F^2
    \le
    \frac{B^2}{K_{\mathrm{eff},s}},
    \quad
    K_{\mathrm{eff},s}
    =
    \left(\sum_k\widetilde p_k^2\right)^{-1}.
    \label{eq:unique_attenuation}
\end{equation}
Hence, the squared magnitude of the aggregated robot-specific component can decay as $O(K_{\mathrm{eff},s}^{-1})$, corresponding to an
$O(K_{\mathrm{eff},s}^{-1/2})$ decay in its Frobenius norm.
More generally,
\begin{equation}
    \left\|\sum_k\widetilde p_kU_k^r\right\|_F^2
    =
    \sum_k\widetilde p_k^2\|U_k^r\|_F^2
    +
    2\sum_{i<j}\widetilde p_i\widetilde p_j
    \langle U_i^r,U_j^r\rangle_F .
    \label{eq:alignment_energy}
\end{equation}
Thus, positive within-cluster alignment preserves coherent update
energy through the cross terms. CUA is designed to promote such
compatibility before aggregation, while SRB subsequently strengthens
weaker spectral directions that remain attenuated.

\section{Experiments}
\label{sec:experiments}

\subsubsection{Datasets and Non-IID Setup}
\label{subsubsec:data_heterogeneity}
We evaluate on three standard LIBERO suites: LIBERO-Spatial, LIBERO-Object, and LIBERO-Goal. Each suite contains 10 language-conditioned manipulation tasks with 50 expert demonstrations per task. After sequence construction, the three suites contain approximately $62{,}250$, $74{,}507$, and $63{,}728$ training sequences, respectively. Each training sample comprises two RGB observations, a language instruction, proprioceptive state, and a 7-D continuous action sequence.

\paragraph{Same-suite task non-IID} For each suite, we construct a separate non-IID federation with $K{=}5$ clients. Following standard FL practice~\cite{li2022federated}, task sequences are partitioned across clients using a Dirichlet distribution with concentration parameter $\alpha$. We use $\alpha{=}0.5$ by default, inducing both task and quantity skew while retaining coverage of the full suite across the federation. The same partition is used for all methods within a suite to ensure controlled comparison. For the client-participation analysis, all demonstrations are repartitioned for each federation size using the  $\alpha{=}0.5$ setting. This setting isolates \emph{intra-suite task heterogeneity}, since all clients operate within the same LIBERO suite but observe different non-IID subsets of its tasks and demonstrations.

\paragraph{Cross-suite task and environment non-IID} To study stronger heterogeneity, we additionally construct federations across different LIBERO suites. For pairwise settings Object-Goal, Object-Spatial, and Goal-Spatial we use $K{=}4$ clients, with two clients from each participating suite. For the three-suite Object-Goal-Spatial setting, we use $K{=}6$ clients, with two clients per suite. For the main cross-suite experiments, we reuse client shards from the original 5-client ($\alpha{=}0.5$) partitions; for the client-participation analysis, all demonstrations are repartitioned for each federation size using the same $\alpha{=}0.5$ setting. Consequently, cross-suite federation introduces heterogeneity at two levels: clients remain non-IID \emph{within} each suite, while clients from different suites additionally differ in task families, objects, environments, and state-action distributions. This setting therefore represents a substantially harder form of task and environment-level non-IID manipulation learning.

\subsection{Implementation Details and Baselines}
\label{subsec:exp_sett_and_baselines}

\paragraph{Policy Architecture}
For all methods, we adopt the LIBERO \textsc{ResNet-T} policy architecture~\cite{liu2023libero}, which uses ResNet visual encoders with FiLM language conditioning and pretrained BERT embeddings, followed by a temporal Transformer and a GMM action head that models a multimodal distribution over continuous actions. We select
\textsc{ResNet-T} as a benchmark-standard backbone, as LIBERO found it to provide strong overall performance across its evaluated policy architectures. Its comparatively compact parameterization is also well suited to FL, where repeatedly communicating large contemporary VLA backbones would impose substantial communication overhead. The same architecture is used for all clients and baselines. For FedDRMan clustering, the temporal Transformer serves as the submodule $\Omega$ for computing client-update similarity, while complete local models are aggregated.

\paragraph{Training and Evaluation Protocol}

Following the standard FL convention of evaluating the learned shared model~\cite{mcmahan2017communication}, we evaluate each federated policy on all $10$ tasks of the corresponding LIBERO suite using closed-loop MuJoCo rollouts with a 250-step horizon. Task success rate (SR) is the primary metric. We evaluate aggregated models every $5$ communication rounds using $10$ rollouts per task for checkpoint selection, and report final performance using 20 randomized rollouts per task. All results are averaged over three seeds.

%Following the standard FL convention of evaluating the learned shared model~\cite{mcmahan2017communication}, we evaluate each federated method's jointly trained policy on all tasks of the corresponding LIBERO suite. Policies are evaluated by closed-loop MuJoCo rollouts over all $10$ tasks, of the corresponding LIBERO suite,  with a 250-step horizon, and task success rate (SR), is used as the primary metric; We checkpoint and evaluate the aggregated policy every $5$ rounds using $10$ rollout trajectories per task for model selection. Aggregated models are evaluated every $5$ rounds using $10$ rollouts/task for checkpoint selection; final evaluation uses $20$ randomized rollouts/task. All SR results are averaged over three seeds.

% Following Robomimic~\cite{robomimic}, we report the checkpoint achieving the highest SR, since lower imitation loss does not necessarily correspond to better closed-loop performance. Final evaluation uses $20$ randomized rollouts per task~\cite{liu2023libero}.All reported SR results are averaged over three random seeds.

\paragraph{Hyperparameters} All FL methods are trained with $1$ local epoch per FL round and a batch size of $32$ for $50$ communication rounds. We use AdamW with a learning rate of $10^{-4}$, weight decay of $10^{-4}$, and gradient clipping at $100$. The same policy architecture, optimizer, client partition, and local training budget are used across federated baselines.
%Use variables   
We use $C=1$ for same-suite task non-IID federation (all clients belong to one group), $C=2$ for pairwise cross-suite settings, and $C=3$ for the three-suite setting. Empirically, for the subspace loss, we use a fixed rank of 48, a distillation weight of 1.0, and feature noise with standard deviation 0.01 for SVD stabilization. Unless specified otherwise, the modality weights are set to 0.03 for language, vision, and proprioception, and 0.005 for action distillation. Suite-specific variants are treated as hyperparameter ablations rather than separate methods. We set \(\beta=0.90\), \(0.85\), and \(0.90\) for same-suite Object, Goal, and Spatial, respectively, and \(\beta=0.90\) for all pairwise cross-suite, and $\beta=0.85$ for three suite settings.

\paragraph{Baselines} Baselines include: 
(i) \textsc{Centralized}, where BC is performed on the pooled training data from all clients; 
(ii) \textsc{Local}, where each client trains independently using only its local data without knowledge sharing; 
(iii) \textsc{FedAvg}~\cite{mcmahan2017communication}, the standard FL baseline that performs sample-weighted parameter averaging of locally trained models; 
(iv) \textsc{FedProx}~\cite{li2020fedprox}, extends FedAvg with a proximal regularization term to limit model drift, using $\mu{=}10^{-3}$; 
(v) \textsc{FedKD}~\cite{FedKD}, a federated generic mutual-distillation framework, adapted to our continuous-action LIBERO policies. We implement the FedKD-A variant (adaptive mutual distillation).
(vi) \textsc{FedVLA}~\cite{miao2025fedvla}, we implement its Dual-Gating mixture-of-experts (MoE) trunk and Expert-Driven Aggregation as specified, and train with our standard GMM negative log-likelihood (NLL) objective rather than the original Huber loss, since Huber-on-the-mean leaves action variance untrained while inference samples from the full distribution.
All methods are evaluated using the same non-IID client partitions and closed-loop evaluation protocol.

\section{Results}
\label{sec:results}

\subsection{Task Non-IID Performance}
\label{subsec:same_suit_perf}

\begin{table*}[!h]
\centering
\caption{Experimental results of same-suite task non-IID($\alpha{=}0.5$) federated evaluation across three different LIBERO task suites. All metrics are measured based on SR(\%). Bold denotes the best
federated result.}
\label{tab:insuite_main_results}
\setlength{\tabcolsep}{4pt}
\resizebox{0.9\textwidth}{!}{%
\begin{tabular}{lcccccccc}
\hline
\multirow{2}{*}{Method}
  & \multicolumn{2}{c}{Object}
  & \multicolumn{2}{c}{Goal}
  & \multicolumn{2}{c}{Spatial}
  & \multirow{2}{*}{\shortstack{Mean\\SR (\%) $\uparrow$}}
  & \multirow{2}{*}{\shortstack{Comm.\\MB / Params}} \\
\cmidrule(lr){2-3}\cmidrule(lr){4-5}\cmidrule(lr){6-7}
  & SR (\%) $\uparrow$ & Best Round
  & SR (\%) $\uparrow$ & Best Round
  & SR (\%) $\uparrow$ & Best Round
  & & \\
\hline

Centralized BC
  & $56.0 \pm 2.0$ & 35
  & $60.4 \pm 2.0$ & 50
  & $46.0 \pm 4.0$ & 50
  & $54.1$ & - / - \\

Local BC
  & $27.2 \pm 2.0$ & 50
  & $39.6 \pm 4.0$ & 50
  & $37.2 \pm 3.0$ & 50
  & $34.7$ & - / - \\
  \hline

FedAvg~\cite{mcmahan2017communication}
  & $75.2 \pm 3.0$ & 50
  & $70.0 \pm 2.0$ & 25
  & $62.0 \pm 3.0$ & 25
  & $69.1$ & 21.53 / 5.38M \\

FedProx~\cite{li2020fedprox} 
  & $70.8 \pm 2.0$ & 40
  & $72.0 \pm 2.0$ & 35
  & $52.0 \pm 4.0$ & 35
  & $64.9$ & 21.53 / 5.38M \\

FedKD~\cite{FedKD}
  & $20.0 \pm 2.0$ & 40
  & $56.0 \pm 4.0$ & 35
  & $44.0 \pm 3.0$ & 25
  & $40.0$ & 21.53 / 5.38M \\

FedVLA~\cite{miao2025fedvla}
  & $60.0\pm 4.0$ & 40
  & $65.0\pm 4.0$ & 40
  & $30.0\pm 3.0$ & 35
  & $51.7$ & 3.70 / 0.925M \\

\rowcolor{green!12}
FedDRMan (ours)
  & $\textbf{88.0}\pm 3.0$ & 40
  & $\textbf{72.0}\pm 3.0$ & 40
  & $\textbf{82.0}\pm 4.0$ & 40
  & $\textbf{80.67}$ & 21.53 / 5.38M \\

\hline
\end{tabular}
}
\vspace{-1.0em}
\end{table*}

Table~\ref{tab:insuite_main_results} compares methods under the same-suite task non-IID setting. In Local BC, no shared policy is learned; thus, all client policies are evaluated on the full suite task set and their average SR is reported. Most collaborative training methods outperform Local BC, demonstrating the benefit of leveraging knowledge across participating clients. FedDRMan achieves the highest SR on LIBERO-Object and LIBERO-Spatial, outperforming the strongest federated baseline by 12.8 pp ($88.0\%$ vs.\ $75.2\%$) and 20.0 pp ($82.0\%$ vs.\ $62.0\%$), respectively. On LIBERO-Goal, FedDRMan reaches $72.0\%$, matching FedProx and exceeding FedAvg by $2.0$ pp. Compared with FedKD, FedDRMan improves Object, Goal, and Spatial by $68.0$, $16.0$, and $38.0$ pp, respectively, suggesting that the evaluated distillation baseline alone is insufficient under strong client heterogeneity. Overall, FedDRMan achieves a mean SR of $80.7\%$, compared with $69.1\%$ for FedAvg, while retaining the same communication payload. FedDRMan attains its best SR at round $40$ across all three suites, within the common 50-round budget. These results highlight the limitations of conventional aggregation and non-IID mitigation strategies even under same-suite task heterogeneity, while supporting the complementary roles of multimodal subspace distillation and structured aggregation.

\subsection{Task and Environment Non-IID Performance}
\label{subsec:cross_suite_perf}

Table~\ref{tab:crosssuite_per_suite} shows that cross-suite federation is substantially more difficult for conventional federated methods than the in-suite setting, with FedAvg, FedProx, and FedKD often attaining very low SR when clients specialize in different suites. In Local BC, no shared policy is learned; thus, each client policy is evaluated on the full task set of its corresponding suite, and the average SR across clients from that suite is reported. In the Object-Goal setting, FedDRMan reaches $85.0\%/70.0\%$, outperforming FedAvg by $67.0/70.0$ pp and FedProx by $69.0/60.0$ pp. A similar trend appears for Object-Spatial, where FedDRMan achieves $95.0\%$ on Object and $45.0\%$ on Spatial, compared with only $18.0\%/0.0\%$ for FedAvg. FedDRMan also outperforms FedVLA in both Object--Goal and Object-Spatial settings. These large gaps suggest that conventional aggregation is particularly vulnerable to incompatible cross-suite updates, which FedDRMan handles more effectively. The Goal-Spatial setting remains more challenging: FedDRMan achieves the strongest Goal performance, while remaining competitive on Spatial. Under the hardest three-suite federation, performance generally decreases, with the largest degradation on Goal, indicating that increasing suite heterogeneity creates stronger interference. Nevertheless, FedDRMan retains markedly higher SR than FedAvg and FedProx, supporting the complementary roles of local subspace distillation, compatibility-aware clustering, and spectral rebalancing for cross-suite knowledge integration.

\begin{table*}[t]
\centering
\caption{Experimental results of cross-suite task and environment non-IID ($\alpha{=}0.5$) federated evaluation across different LIBERO suite combinations. All metrics are measured based on SR(\%). Bold denotes the best federated result.}

\label{tab:crosssuite_per_suite}
\setlength{\tabcolsep}{4pt}
\resizebox{\textwidth}{!}{%
\begin{tabular}{lcccccccccc}
\hline
\multirow{2}{*}{Method}
  & \multicolumn{2}{c}{Object-Goal}
  & \multicolumn{2}{c}{Object-Spatial}
  & \multicolumn{2}{c}{Goal-Spatial}
  & \multicolumn{3}{c}{Object-Goal-Spatial}
  & \multirow{2}{*}{\shortstack{Mean\\SR (\%) $\uparrow$}} \\ 
\cmidrule(lr){2-3}
\cmidrule(lr){4-5}
\cmidrule(lr){6-7}
\cmidrule(lr){8-10}
  & Obj & Goal
  & Obj & Spat
  & Goal & Spat
  & Obj & Goal & Spat
  & \\
\hline

Centralized BC
  & $78.0{\pm}2.0$ & $78.0{\pm}2.0$
  & $62.0{\pm}1.0$ & $46.0{\pm}2.0$
  & $58.0{\pm}2.0$ & $54.0{\pm}3.0$
  & $44.0{\pm}4.0$ & $60.0{\pm}3.0$ & $42.0{\pm}3.0$
  & 58.0 \\

Local BC
  & $15.0{\pm}2.0$ & $48.0{\pm}3.0$
  & $15.0{\pm}2.0$ & $35.0{\pm}2.0$
  & $48.0{\pm}3.0$ & $35.0{\pm}2.0$
  & $15.0{\pm}2.0$ & $48.0{\pm}3.0$ &$35.0{\pm}2.0$
  & 32.7 \\
  \hline

FedAvg~\cite{mcmahan2017communication}
  & $18.0{\pm}2.0$ & $0.0{\pm}0.0$
  & $18.0{\pm}3.0$ & $0.0{\pm}0.0$
  & $12.0{\pm}1.0$ & $12.0{\pm}0.0$
  & $2.0{\pm}1.0$ & $14.0{\pm}3.0$ & $8.0{\pm}2.0$
  & 9.3 \\

FedProx~\cite{li2020fedprox}
  & $16.0{\pm}1.0$ & $10.0{\pm}2.0$
  & $16.0{\pm}1.0$ & $2.0{\pm}1.0$
  & $12.0{\pm}2.0$ & $0.0{\pm}0.0$
  & $4.0{\pm}2.0$ & $0.0{\pm}0.0$ & $2.0{\pm}1.0$
  & 6.9 \\

FedKD~\cite{FedKD}
  & $20.0{\pm}2.0$ & $8.0{\pm}2.0$
  & $4.0{\pm}2.0$ & $6.0{\pm}2.0$
  & $14.0{\pm}3.0$ & $14.0{\pm}4.0$
  & $0.0{\pm}0.0$ & $2.0{\pm}0.0$ & $0.0{\pm}0.0$
  & $7.6$ \\

FedVLA~\cite{miao2025fedvla}
  & $35.0{\pm}3.0$ & $60.0{\pm}5.0$
  & $65.0{\pm}2.0$ & $30.0{\pm}3.0$
  & $40.0{\pm}4.0$ & $\textbf{60.0}{\pm}3.0$
  & $55.0{\pm}3.0$ & $\textbf{40.0}{\pm}4.0$ & $35.0{\pm}3.0$
  &$46.7$ \\

\rowcolor{green!12} FedDRMan (ours)
  & $\textbf{85.0}{\pm}2.0$ & $\textbf{70.0}{\pm}3.0$
  & $\textbf{95.0}{\pm}4.0$ & $\textbf{45.0}{\pm}4.0$
  & $\textbf{65.0}{\pm}3.0$ & $55.0{\pm}3.0$
  & $\textbf{75.0}{\pm}4.0$ & $35.0{\pm}4.0$ & $\textbf{45.0}{\pm}5.0$
  & \textbf{63.3} \\

\hline
\end{tabular}%
}
\vspace{-1.0em}
\end{table*}

\subsection{Task Specific Success}
\label{subsec:task_specific_success}

\begin{figure}[!bh]
    \centering

    \begin{subfigure}[t]{0.99\columnwidth}
        \centering
        \includegraphics[width=\linewidth]{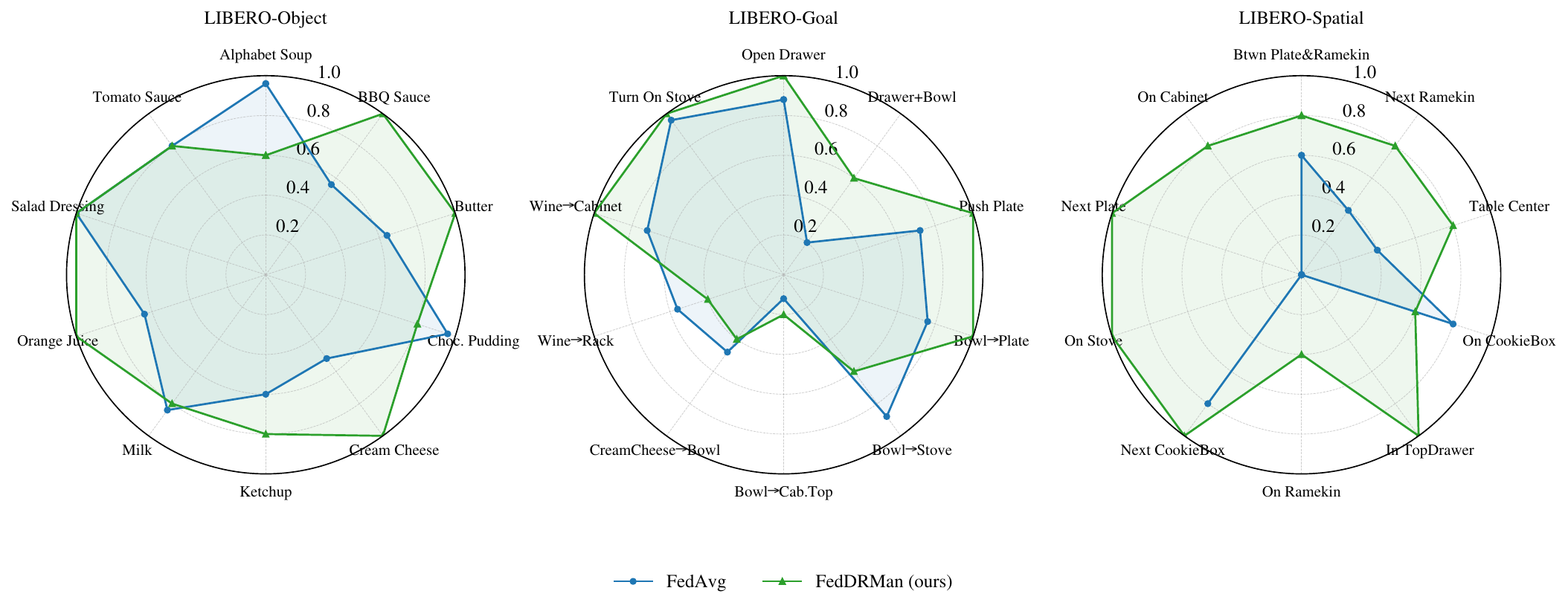}
        % \caption{}
        \label{fig:same_object}
    \end{subfigure}

    \vspace{-0.4em}

    \caption{
    Task sensitivity analysis across different suites.
    }
    \label{fig:task_sensitivity}
    \vspace{-1.0em}
\end{figure}

Figure~\ref{fig:task_sensitivity} further compares the per-task performance, with success rates normalized to $[0,1]$ for visualization. We compare against FedAvg, the strongest competing federated baseline in the same-suite setting, as cross-suite gains are already substantially larger. FedDRMan outperforms or matches FedAvg on 23 of the 30 tasks across the three suites. On LIBERO-Object, notable gains are observed for \emph{Cream Cheese} ($+48$ pp), \emph{Butter} ($+36$ pp), and \emph{Ketchup} ($+20$ pp) relative to FedAvg. On LIBERO-Goal, FedDRMan improves \emph{Bowl on Plate} by $24$ pp, \emph{Wine Bottle on Cabinet} by $28$ pp, and \emph{Push Plate} by $28$ pp, although FedAvg remains stronger on a few tasks. The advantage is most pronounced on LIBERO-Spatial, where FedDRMan raises both \emph{On Stove} and \emph{Next to Plate} from $0\%$ to $100\%$, while improving \emph{In Top Drawer} from FedAvg's $0\%$ to $100\%$ ($+100$ pp). These results indicate that FedDRMan particularly improves task-specific manipulation behaviours that are poorly preserved by conventional federated aggregation.

\vspace{-0.4em}

\subsection{Non-IID Sensitivity}
\label{subsec:alpha_sensitivity}

\begin{figure}[!bh]
    \centering

    \begin{subfigure}[t]{0.48\columnwidth}
        \centering
        \includegraphics[width=\linewidth]{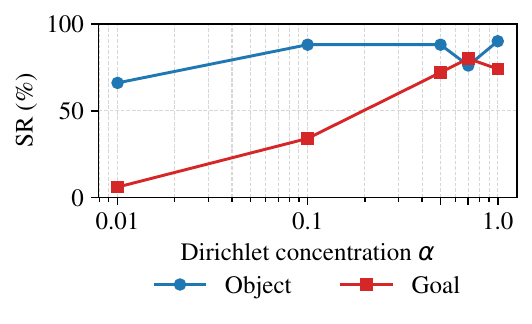}
        \caption{Task non-IID.}
        \label{fig:noniid_same_o_g}
    \end{subfigure}
    \hfill
    \begin{subfigure}[t]{0.48\columnwidth}
        \centering
        \includegraphics[width=\linewidth]{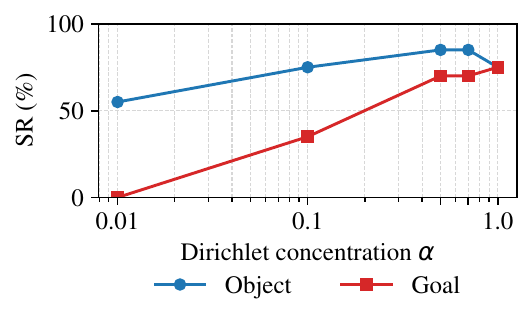}
        \caption{Task and environment non-IID.}
        \label{fig:noniid_cross_o_g}
    \end{subfigure}

    \caption{Sensitivity of FedDRMan to varying Dirichlet non-IID severity ($\alpha$).}
    
    \label{fig:noniid_sensitivity}
    \vspace{-1.0em}
\end{figure}

We further evaluate FedDRMan under different degrees of client heterogeneity by varying the Dirichlet concentration parameter $\alpha$, where smaller values indicate stronger non-IID partitions. As shown in Fig.~\ref{fig:noniid_sensitivity}, performance generally improves as $\alpha$ increases. In the cross-suite Object--Goal setting, Object SR rises from roughly $55\%$ at $\alpha{=}0.01$ to the $75$--$85\%$ range for $\alpha{\geq}0.1$, while Goal improves substantially, from near $0\%$ at $\alpha{=}0.01$ to approximately $70$--$75\%$ for less heterogeneous settings. A similar trend appears in the same-suite task non-IID setting, where both Object and Goal improve as heterogeneity decreases, although not monotonically at larger $\alpha$; Goal is particularly sensitive to severe non-IIDness, with SR dropping to nearly $0\%$ at $\alpha{=}0.01$.

The trend is not strictly monotonic; for example, Object exhibits a small drop at larger $\alpha$ in some runs. Nevertheless, the overall pattern is consistent across the sweeps: the most heterogeneous partitions lead to markedly lower success rates, whereas moderate-to-high $\alpha$ values yield substantially stronger performance. This confirms that severe statistical heterogeneity remains a major challenge for federated manipulation and highlights the importance of FedDRMan's representation-preserving and clustered aggregation mechanisms under strongly non-IID conditions.

\subsection{Scalability}
\label{subsec:scalability_communication_analysis}

Performance is generally strongest at moderate client counts
($k{=}6$--$10$) and degrades when scaling to $k{=}20$.
Under fixed Dirichlet($\alpha{=}0.5$) partitioning, the total demonstration set is divided across $k$ clients, so increasing $k$ reduces the amount of local data available per client. At $k{=}20$, each client therefore receives a substantially smaller and potentially noisier local shard, which can reduce the quality of local updates before aggregation. 

Goal is noticeably more sensitive to this effect than Object: from $k{=}10$ to $k{=}20$, Goal drops from $80\%$ to $58\%$ in the same-suite setting and from $70\%$ to $60\%$ cross-suite, whereas Object decreases more moderately from $85\%$ to $75\%$ and from $89\%$ to $82\%$, respectively. 
Goal is more sensitive than Object to increasing client count, suggesting greater sensitivity to reduced per-client demonstration coverage (from $k{=}10$ to $k{=}20$). Goal drops from $80\%$ to $58\%$ in the same-suite setting and from $70\%$ to $60\%$ cross-suite, whereas Object decreases more moderately from $85\%$ to $75\%$ and from $89\%$ to $82\%$, respectively. Performance at smaller client counts ($k$) is also slightly below the best intermediate configurations, indicating a trade-off between the amount of local data available to each client and the diversity of updates available for federation. This suggests that, although some suites and tasks are more sensitive to reduced per-client demonstration coverage, FedDRMan remains effective across a broad range of client counts.

\begin{figure}[!t]
    \centering

    \begin{subfigure}[t]{0.48\columnwidth}
        \centering
        \includegraphics[width=\linewidth]{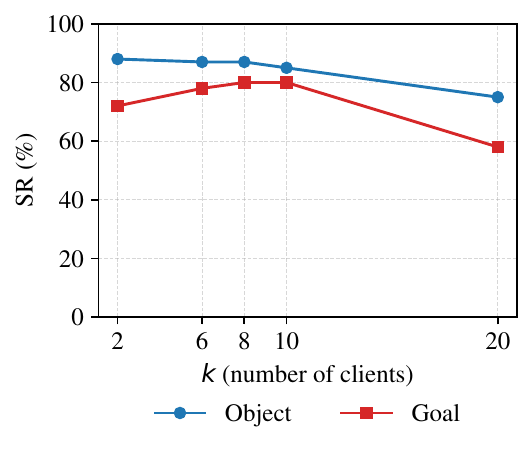}
        \caption{Same-suite scalability.}
        \label{fig:noniid_same_object}
    \end{subfigure}
    \hfill
    \begin{subfigure}[t]{0.48\columnwidth}
        \centering
        \includegraphics[width=\linewidth]{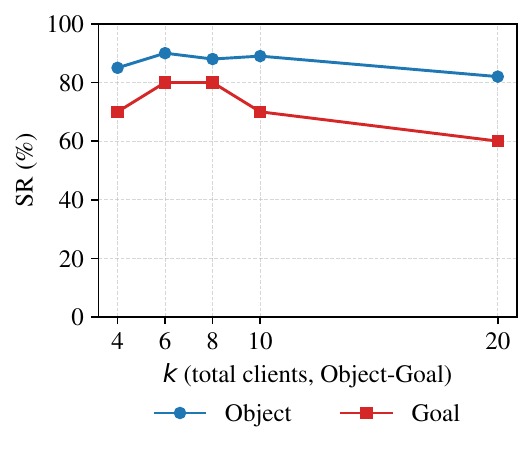}
        \caption{Cross-suite scalability }
        \label{fig:noniid_same_goal}
    \end{subfigure}

    \vspace{-0.2em}

    \caption{
    Analysis of sensitivity to client participation variation.
    }
    \label{fig:scalability_and_communication}
    \vspace{-1.0em}
\end{figure}

\subsection{Ablations}
\label{subsec:distillation_ablations}

Table~\ref{tab:ablation_compact} isolates the contribution of each FedDRMan component in the Object-Goal cross-suite setting. Using subspace alignment alone reduces SR from $70.0/60.0$ to $60.0/50.0$,
indicating that preserving representation geometry without explicitly preserving the teacher's action distribution can over-constrain local adaptation while still allowing policy-level behavior to drift.
Conversely, using action distillation alone also degrades performance, particularly on Object, where SR drops to $35.0\%$ (and to $55.0\%$ on Goal). This suggests that action policy guidance alone is insufficient under strong cross-suite heterogeneity, since it does not explicitly constrain drift in the visual, language, and proprioceptive representations that support manipulation. When subspace alignment and
action distillation are combined, however, performance substantially recovers, reaching $80.0\%$ on Object and $60.0\%$ on Goal. This supports their complementary roles: subspace alignment preserves transferable multimodal representation structure, while action
distillation anchors the resulting control behavior. SRB further improves both suites to $85.0\%$ and $70.0\%$, respectively, yielding the strongest overall configuration. 
Most importantly, removing CUA causes complete failure in this setting ($0.0\%$ on both suites), falling below plain FedAvg. This indicates that the proposed distillation and spectral rebalancing are ineffective when incompatible cross-suite updates are aggregated into a single server model, highlighting the importance of compatibility-aware grouping under strong heterogeneity.

This result is also consistent with the aggregation analysis in Sec.~\ref{subsubsec:heterogeneous_analysis}, without CUA, poorly aligned client-specific components can cancel during aggregation, whereas compatibility-based grouping
preserves more coherent update energy. Overall, the results suggest that local representation and policy-level preservation and server-side compatible update aggregation address complementary sources of non-IID interference.

\begin{table}[t]
\centering
\caption{
Component-wise ablation of FedDRMan on the Object-Goal cross-suite task and environment non-IID setting. 
}
\label{tab:ablation_compact}
\setlength{\tabcolsep}{4pt}
\resizebox{0.9\columnwidth}{!}{%
\begin{tabular}{ccccccc}
\hline
$\mathcal{L}_{\mathrm{BC}}$
& $\mathcal{L}_{\mathrm{sub}}$
& $\mathcal{L}_{\mathrm{action}}$
& SRB
& CUA
& SR (Object) $\uparrow$
& SR (Goal) $\uparrow$ \\
\hline

$\checkmark$
& $\times$
& $\times$
& $\times$
& $\checkmark$
& 70.0
& 60.0 \\

$\checkmark$
& $\times$
& $\checkmark$
& $\times$
& $\checkmark$
& 35.0
& 55.0 \\

$\checkmark$
& $\checkmark$
& $\times$
& $\times$
& $\checkmark$
& 60.0
& 50.0 \\

$\checkmark$
& $\checkmark$
& $\checkmark$
& $\times$
& $\checkmark$
& 80.0
& 60.0 \\

$\checkmark$
& $\checkmark$
& $\checkmark$
& $\checkmark$
& $\checkmark$
& \textbf{85.0}
& \textbf{70.0} \\

$\checkmark$
& $\checkmark$
& $\checkmark$
& $\checkmark$
& $\times$
& 0.0
& 0.0 \\

\hline
\end{tabular}%
}
\vspace{-1.0em}
\end{table}

\subsection{CUA Effectiveness}
\label{subsec:aggregation_effectiveness}

\begin{figure}[!tbh]
    \centering

    \begin{subfigure}[t]{0.48\columnwidth}
        \centering
        \includegraphics[width=\linewidth]{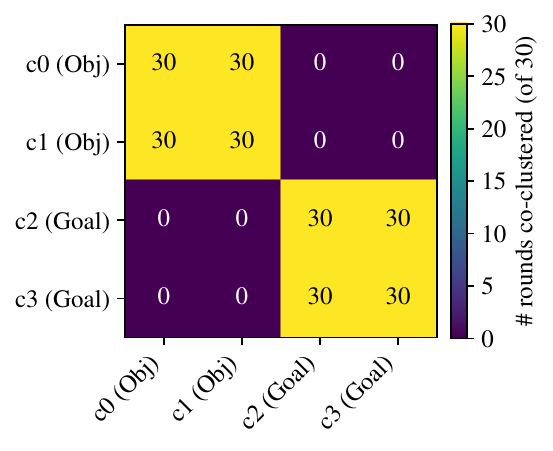}
        \caption{Cross-suite aggregation.}
        \label{fig:cross_suite_agg_heatmap}
    \end{subfigure}
    \hfill
    \begin{subfigure}[t]{0.48\columnwidth}
        \centering
        \includegraphics[width=\linewidth]{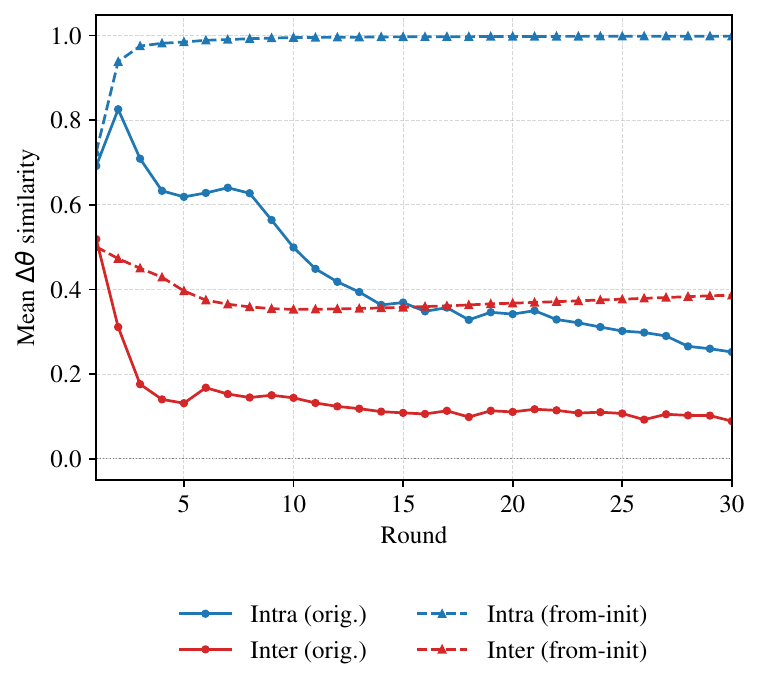}
        \caption{Cross-suite cluster distance.}
        \label{fig:cross_suit_cluster_dist}
    \end{subfigure}

    \vspace{0.4em}

    \caption{
    Aggregation effectiveness.
    }
    \label{fig:aggregation_effectiveness}
    % \vspace{-1.0em}
\end{figure}

\paragraph{Compatible Grouping} We further analyze whether compatibility-aware clustering successfully separates clients with compatible local updates. Fig.~\ref{fig:cross_suite_agg_heatmap} shows that the two Object clients are co-clustered in all 30 rounds (best performance round), and the two Goal clients exhibit the same behavior, while no Object-Goal client pair is ever assigned to the same cluster. This indicates that the local update direction provides a stable signal of suite-level compatibility rather than producing transient round-wise groupings.

\paragraph{Update Dynamics}
Fig.~\ref{fig:cross_suit_cluster_dist} analyzes the update geometry used for client clustering. The clustering signal, computed from each client's accumulated update from initialization (triangle curves), exhibits a persistent separation between intra- and inter-cluster similarity. Intra-cluster similarity rapidly approaches $1.0$ and remains nearly constant, indicating highly consistent directions within each cluster. In contrast, inter-cluster similarity decreases from approximately $0.5$
to $0.35$--$0.4$ and remains substantially lower, confirming distinct update directions across heterogeneous suites.

We also report the similarity of incremental per-round updates
(circle curves). Both intra- and inter-cluster similarities decrease as local models specialize, while intra-cluster similarity remains consistently higher. These curves are diagnostic only; clustering uses the accumulated from-initial updates. Together, the two views indicate that FedDRMan identifies coherent robot-policy groups despite differing short-term optimization directions, reducing destructive cross-suite averaging before SRB preserves weaker directions within each compatible
aggregate. This separation is consistent with the analysis in Sec.~\ref{subsubsec:heterogeneous_analysis} higher within-cluster
alignment increases the positive cross terms in Eq.~\eqref{eq:alignment_energy}, reducing attenuation of coherent
robot-specific update directions.

\section{Conclusion}
\label{sec:conclusion}

FedDRMan addresses two coupled challenges in non-IID federated robotic manipulation: multimodal policy drift during local adaptation and destructive interference among heterogeneous robot-policy updates.
It combines server-guided subspace and action distillation with compatibility-aware clustering and spectral rebalancing to preserve transferable policy structure throughout local and server-side learning. Across LIBERO Object, Goal, and Spatial, FedDRMan achieves $80.7\%$ mean SR, improving over the strongest evaluated federated baseline by $11.6$ pp with no additional communication payload. Cross-suite, heterogeneity-sensitivity, and client-scaling experiments further show its robustness under increasingly heterogeneous federations, while ablations confirm the complementary roles of local distillation, CUA, and SRB. Future work will investigate partial client participation, longer-horizon and more compositional manipulation tasks and real-robot deployments.

\balance{}

\bibliographystyle{IEEEtran}  
\bibliography{references}

\end{document}